\documentclass[runningheads]{llncs}
\usepackage{booktabs}
\usepackage{multirow}
\usepackage{amsmath}
\usepackage{amssymb}
\usepackage{amsfonts}
\usepackage{makecell}
\usepackage[T1]{fontenc}
\usepackage{bm}
\usepackage{cite}
\usepackage{array}
\usepackage{url}
\usepackage[misc]{ifsym}
\usepackage{bbding}
\usepackage[colorlinks,
            linkcolor=blue,
            anchorcolor=blue,
            citecolor=blue]{hyperref}

\usepackage{graphicx}
\begin{document}
\title{ModeT: Learning Deformable Image Registration via Motion Decomposition Transformer}
\titlerunning{Motion Decomposition Transformer for Registration}
%
\author{Haiqiao Wang \and Dong Ni \and Yi Wang\Envelope}


%
\authorrunning{H. Wang et al.}
%
\institute{
	Smart Medical Imaging, Learning and Engineering (SMILE) Lab,\\
	Medical UltraSound Image Computing (MUSIC) Lab,\\
	School of Biomedical Engineering,
	Shenzhen University Medical School,\\
	Shenzhen University, Shenzhen, China\\
	\email{onewang@szu.edu.cn} \\
}
%
\maketitle              
\begin{abstract}
The Transformer structures have been widely used in computer vision and have recently made an impact in the area of medical image registration.
However, the use of Transformer in most registration networks is straightforward.
These networks often merely use the attention mechanism to boost the feature learning as the segmentation networks do, but do not sufficiently design to be adapted for the registration task.
In this paper, we propose a novel motion decomposition Transformer (ModeT) to explicitly model multiple motion modalities by fully exploiting the intrinsic capability of the Transformer structure for deformation estimation.
The proposed ModeT naturally transforms the multi-head neighborhood attention relationship into the multi-coordinate relationship to model multiple motion modes.
Then the competitive weighting module (CWM) fuses multiple deformation sub-fields to generate the resulting deformation field.
Extensive experiments on two public brain magnetic resonance imaging (MRI) datasets show that our method outperforms current state-of-the-art registration networks and Transformers, demonstrating the potential of our ModeT for the challenging non-rigid deformation estimation problem.
\textit{The benchmarks and our code are publicly available at} \url{https://github.com/ZAX130/SmileCode}.

\end{abstract}
\section{Introduction}
Deformable image registration has always been an important focus in the society of medical imaging, which is essential for the preoperative planning, intraoperative information fusion, disease diagnosis and follow-ups~\cite{sotiras2013deformable, fu2020deep}.
The deformable registration is to solve the non-rigid deformation field to warp the moving image, so that the warped image can be anatomically similar to the fixed image.
Let $ I_f, I_m \in\mathbb{R}^{H\times W\times L} $ be the fixed and moving images ($H, W, L$ denote image size), in the deep-learning-based registration paradigm, it is often necessary to employ a spatial transformer network (STN)~\cite{jaderberg2015spatial} to apply the estimated sampling grid $ G\in\mathbb{R}^{H \times W \times L \times 3}$ to the moving image, where $G$ is obtained by adding the regular grid and the deformation field.
For any position $p\in\mathbb{R}^3$ in the sampling grid, $G(p)$ represents the corresponding relation,  which means that the voxel at position $p$ in the fixed image corresponds to the voxel at position $G(p)$ in the moving image.
That is to say, image registration can be understood as finding the corresponding voxels between the moving and fixed images, and converting this into the relative positional relationship between voxels, which is very similar to the calculation method of Transformer~\cite{dosovitskiyimage}.

Transformers have been successfully used in the society of computer vision and have recently made an impact in the field of medical image computing~\cite{li2023transforming, He2022}.
In medical image registration, there are also several related studies that employ Transformers to enhance network structures to obtain better registration performance, such as Transmorph~\cite{Chen2022a}, Swin-VoxelMorph~\cite{Zhu2022}, Vit-V-Net~\cite{Chen2021}, etc.
The use of Transformer in these networks, however, often merely leverages the self-attention mechanism in Transformers to boost the feature learning (the same as the segmentation tasks do), but does not sufficiently design for the registration tasks.
Some other methods use cross-attention to model the corresponding relationship between moving and fixed images, such as Attention-Reg~\cite{Song2021} and Xmorpher~\cite{shi2022xmorpher}.
The cross-attention Transformer (CAT) module is used in the bottom layer of Attention-Reg~\cite{Song2021} and each layer in Xmorpher~\cite{shi2022xmorpher} to establish the relationship between the features of moving and fixed images.
However, the usage of Transformer in~\cite{shi2022xmorpher, Song2021} is still limited to improving the feature learning, with no additional consideration given to the relationship between the attention mechanism and the deformation estimation.
Furthermore, due to the large network structure of~\cite{shi2022xmorpher}, only small windows can be created for similarity calculation, which may result in performance degradation.
Few studies consider the relationship between attention and deformation estimation, such as Coordinate Translator~\cite{Liu2022} and Deformer~\cite{Chen2022}.
Deformer~\cite{Chen2022} uses the calculation mode of multiplication of attention map and Value matrix in transformer to weight the predicted basis to generate the deformation field, but its attention map calculation is only the concatenation and projection of moving and fixed feature maps, without using similarity calculation part.
Coordinate Translator~\cite{Liu2022} calculates the matching score of the fixed feature map and the moving feature map.
Then the computed scores are employed to re-weight the deformation field.
However, for feature maps with coarse-level resolution, a voxel often has multiple possibilities of different motion modes~\cite{Zheng2022}, which is not considered in~\cite{Liu2022}.

In this study, we propose a novel motion decomposition transformer (ModeT) to explicitly model multiple motion modalities by fully exploiting the intrinsic capability of the Transformer structure for deformation estimation.
Experiments on two public brain magnetic resonance imaging (MRI) datasets demonstrate our method outcompetes several cutting-edge registration networks and Transformers.
The main contributions of our work are summarized as follows:
\begin{itemize}
	\item[$\bullet$] We propose to leverage the Transformer structure to naturally model the correspondence between images and convert it into the deformation field, thus explicitly separating the two tasks of feature extraction and deformation estimation in deep-learning-based registration networks in which to make the registration procedure more sensible.
	
	\item[$\bullet$] The proposed ModeT makes full use of the multi-head neighborhood attention mechanism to efficiently model multiple motion modalities, and then the competitive weighting module (CWM) fuses multiple deformation sub-fields in a competitive way, which can improve the interpretability and consistency of the resulting deformation field.

	\item[$\bullet$] The pyramid structure is employed for feature extraction and deformation propagation, and is beneficial to reduce the scope of attention calculation required for each level.
\end{itemize}

\section{Method}
\subsection{Network Overview}
\begin{figure}[t]
	\centering
	\includegraphics[width=1.0\columnwidth]{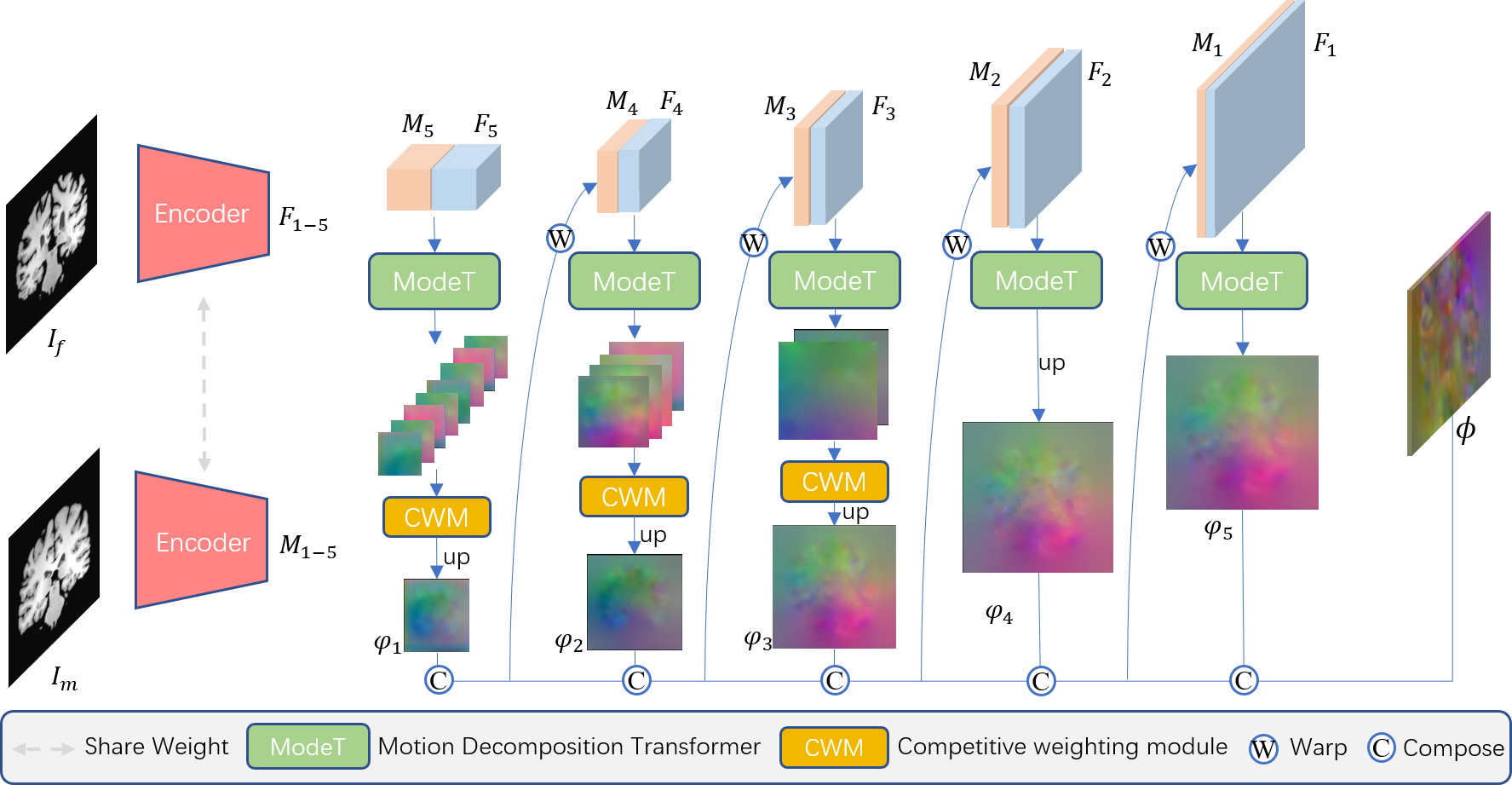}
	\caption{Illustration of the proposed deformable registration network. The encoder takes the fixed image $I_f$ and moving image $I_m$ as input to extract hierarchical features $F_1$-$F_5$ and $M_1$-$M_5$. The motion decomposition transformer (ModeT) is used to generate multiple deformation sub-fields and the competitive weighting module (CWM) fuses them. Finally the decoding pyramid outputs the total deformation field $\phi$.}
	\label{f:overall}
\end{figure}

The proposed deformable registration network is illustrated in Fig.~\ref{f:overall}.
We employ a pyramidal registration structure, which has the advantage of reducing the scope of attention calculation required at each decoding level and therefore alleviating the computational consumption.
Given the fixed image $I_f$ and moving image $I_m$ as input, the encoder extracts hierarchical features using a 5-layer convolutional block, which doubles the number of channels in each layer.
This generates two sets of feature maps $F_1, F_2, F_3, F_4, F_5$ and $M_1 $, $ M_2 $, $ M_3 $, $ M_4 $, $ M_5$.
The feature maps $M_5$ and $F_5$ are sent into the ModeT to generate multiple deformation sub-fields, and then the generated deformation sub-fields are input into the CWM to obtain the fused deformation field $\varphi_1$ of the coarsest decoding layer as the initial of the total deformation field $\phi$.
The moving feature map $M_4$ is deformed using $\phi$, and the deformed moving feature map is fed into the ModeT along with $F_4$ to generate multiple sub-fields, which are input into the CWM to get $\varphi_2$.
Then $\varphi_2$ is compounded with previous total deformation field to generate the updated $\phi$.
The feature maps $M_3$ and $F_3$ go through the similar operations.
As the decoding feature maps become finer, the number of motion modes at position $p$ decreases, along with the number of attention heads we need to model.
At the $F_2/M_2$ and $F_1/M_1$ levels, we no longer generate multiple deformation sub-fields, i.e., the number of attention heads in ModeT is 1.
Finally, the obtained total deformation field $\phi$ is used to warp $I_m$ to obtain the registered image.

To guide the network training, the normalized cross correlation $\mathcal{L}_{\text{ncc}}$~\cite{rao2014application} and the deformation regularization $\mathcal{L}_{\text{reg}}$~\cite{VoxelMorph} is used:
\begin{equation}
\mathcal{L}_{\text{train}} = \mathcal{L}_{\text{ncc}}(I_f, I_m\circ\phi) + \lambda \mathcal{L}_{\text{reg}}(\phi),
\label{e:loss}
\end{equation}
where $\circ$ is the warping operation, and $\lambda$ is the weight of the regularization term.

\subsection{Motion Decomposition Transformer (ModeT)}
\begin{figure}[t]
	\centering
	\includegraphics[width=0.95\columnwidth]{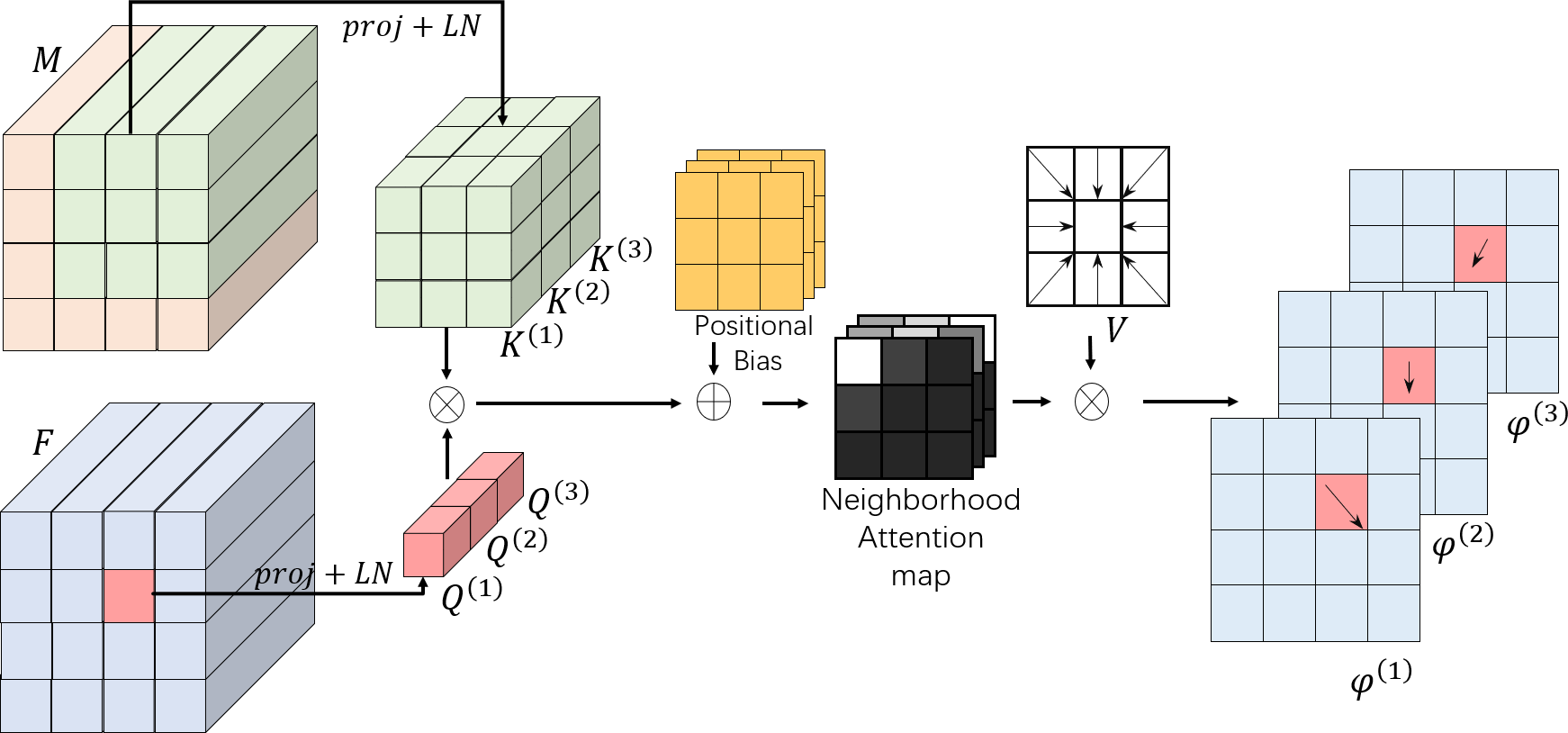}
	\caption{Illustration of the proposed motion decomposition transformer, which employs the multi-head neighborhood attention mechanism to decompose different motion modalities. ($S=3$ in this illustration)}
	\label{f:mcotr}
\end{figure}

In deep-learning-based registration networks, a position $p$ in the low-resolution feature map contains semantic information of a large area in the original image and therefore may often have multiple possibilities of different motion modalities.
To model these possibilities, we employ a multi-head neighborhood attention mechanism to decompose different motion modalities at low-resolution level.
The illustration of the motion decomposition is shown in Fig.~\ref{f:mcotr}.

Let $F, M\in \mathbb{R}^{c\times h\times w\times l}$ stand for the fixed and moving feature maps from a specific level of the hierarchical encoder, where $h, w, l$ denote feature map size and $c$ is the channel number.
The feature maps $F$ and $M$ go through linear projection ($proj$) and LayerNorm ($LN$)~\cite{ba2016layer} to get $Q$ ($query$) and $K$ ($key$):
\begin{equation}
\begin{split}
Q = LN(&proj(F)), \quad	K= LN(proj(M)), \\
Q=&\{Q^{(1)}, Q^{(2)},\dots,Q^{(S)}\},\\
K=&\{K^{(1)}, K^{(2)},\dots,K^{(S)}\},
\end{split}
\end{equation}
where the projection operation is shared weight, and the weight initialization is sampled from $N(0,1e^{-5})$, the bias is initialized to 0.
The $Q$ and $K$ are then divided according to channels, and $S$ represents the number of divided heads.

We then calculate the neighborhood attention map.
We use $c(p)$ to denote the neighborhood of voxel $p$.
For a neighborhood of size $ n\times n\times n$, $||c(p)||=n^3$.
The neighborhood attention map of multiple heads is obtained by:
\begin{equation}
	NA(p,s) = softmax(Q_p^{(s)}\cdot K^{(s)T}_{c(p)}+B^{(s)}),
	\label{e:na}
\end{equation}
where $B \in\mathbb{R}^{S\times n\times n \times n}$ is a learnable relative positional bias, initialized to all zeros.
We pad the moving feature map with zeros to calculate boundary voxels because the registration task sometimes requires voxels outside the field-of-view to be warped.
Equation (\ref{e:na}) shows how the neighborhood attention is computed for the $s$-th head at position $p$, so that the semantic information of voxels on low resolution can be decomposed to compute similarity one by one, in preparation for modeling different motion modalities.
Moreover, the neighborhood attention operation narrows the scope of attention calculation to reduce the computational effort, which is very friendly to volumetric processing.

The next step is to obtain the multiple sub-fields at this level by computing the regular displacement field weighted via the neighborhood attention map:
\begin{equation}
\varphi_p^{(s)} = NA(p,s)V,
\label{e:subsubflow}
\end{equation}
where $\varphi^{(s)}\in\mathbb{R}^{h\times w \times l\times 3}$, $V\in\mathbb{R}^{n\times n \times n}$,
and $V$ ($value$) represents the relative position coordinates for the neighborhood centroid, which is not learned so that the multi-head attention relationship can be naturally transformed into a multi-coordinate relationship.
With the above steps, we obtain a series of deformation sub-fields for this level:
\begin{equation}
	\varphi^{(1)},\varphi^{(2)},\dots,\varphi^{(S)}
	\label{e:warps}
\end{equation}

\subsection{Competitive Weighting Module (CWM)}
\begin{figure}[t]
	\centering
	\includegraphics[width=0.95\columnwidth]{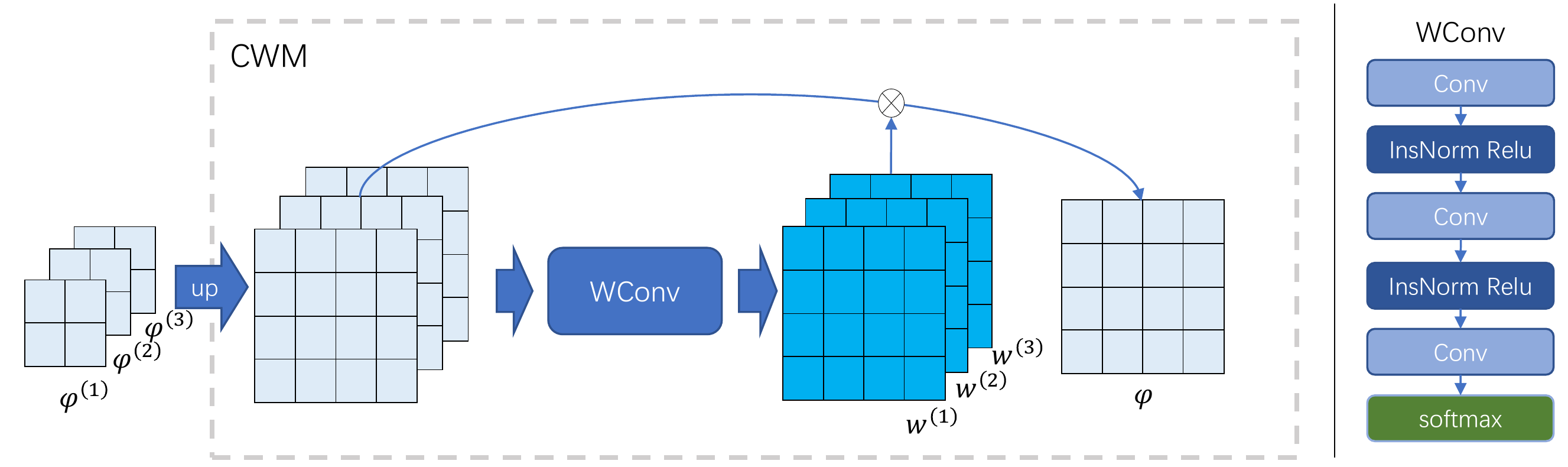}
	\caption{Illustration of the proposed competitive weighting module (CWM).}
	\label{f:cwm}
\end{figure}

Multiple low-resolution deformation fields need to be reasonably fused when deforming a high-resolution feature map.
As shown in Fig.~\ref{f:cwm}, 
we first upsample these deformation sub-fields, then convolve them in three layers to get the score of each sub-field, and use softmax to compete the motion modality for each voxel.
The convolution uses $3\times3\times3$ convolution rather than direct projection because deformation fields often require correlation of adjacent displacements to determine if they are reasonable.
We formulate above competitive weighting operation to obtain the deformation field $\varphi$ at this level as follows:
\begin{equation}
\begin{split}
w^{(1)},w^{(2)},&\dots,w^{(S)}= WConv(cat(\varphi^{(1)},\varphi^{(2)},\dots,\varphi^{(S)})), \\
\varphi &= w^{(1)}\varphi^{(1)} + w^{(2)}\varphi^{(2)}+ ,\dots, +w^{(S)}\varphi^{(S)},
\end{split}
\end{equation}
where $w^{(s)}\in\mathbb{R}^{h\times w \times l}$,
and $\varphi^{(s)}$ has already been upsampled.
$WConv$ represents the ConvBlock used to calculate weights, as shown in the right part of Fig.~\ref{f:cwm}.

\section{Experiments}
\subsubsection{Datasets.}
Experiments were carried on two public brain MRI datasets, including LPBA~\cite{Shattuck2008} and Mindboggle~\cite{Klein2012}.
For LPBA, each MRI volume contains 54 manually labeled region-of-interests (ROIs).
All volumes in LPBA were rigidly pre-aligned to mni305.
30 volumes ($30\times29$ pairs) were employed for training and 10 volumes ($10\times9$ pairs) were used for testing.
For Mindboggle, each volume contains 62 manually labeled ROIs.
All volumes in Mindboggle were affinely aligned to mni152.
42 volumes ($42\times 41$ pairs from the NKI-RS-22 and NKI-TRT-20 subsets) were employed for training, and 20 volumes from OASIS-TRT-20 ($20\times19$ pairs) were used for testing.
All volumes were pre-processed by min-max normalization, and skull-stripping using FreeSurfer~\cite{Fischl2012}.
The final size of each volume was $160\times192\times160$ after a center-cropping operation.

\subsubsection{Evaluation Metrics.}
To quantitatively evaluate the registration performance, Dice score (DSC)~\cite{Dice1945} was calculated as the primary similarity metric to evaluate the degree of overlap between corresponding regions.
In addition, the average symmetric surface distance (ASSD)~\cite{Taha2015} was evaluated, which can reflect the similarity of the region contours.
The quality of the predicted deformation $\phi$ was assessed by the percentage of voxels with non-positive Jacobian determinant (i.e., folded voxels).
All above metrics were calculated in 3D.
A better registration shall have larger DSC, and smaller ASSD and Jacobian.

\begin{table}[t]
	\centering
	\caption{The numerical results of different registration methods on two datasets.}
	\label{tab1}
	\begin{tabular}{lccc|ccc}
		\toprule
		&\multicolumn{3}{c}{Mindboggle (62 ROIs)}&\multicolumn{3}{c}{LPBA (54 ROIs)}\\
		\cmidrule(l{2pt}r{2pt}){2-7}
		& DSC ($\%$) & ASSD & $ \% |J_{\phi}|\le0 $& DSC ($\%$) & ASSD & $ \% |J_{\phi}|\le0 $ \\
		\cmidrule(l{2pt}r{2pt}){1-7}
		\texttt{SyN}~\cite{AVANTS2008} & $56.7\pm1.5$ & $1.38\pm0.09$ & $\textless0.00001\%$ & $70.1\pm6.2$ & $1.72\pm0.12$ & $\textless0.0004\%$  \\
		\texttt{VM}~\cite{VoxelMorph} & $56.0\pm1.6$ & $1.49\pm0.11$ &  $\textless1\%$ & $64.3\pm3.2$&$2.03\pm0.21$& $\textless0.7\%$   \\
		\texttt{TM}~\cite{Chen2022a}  & $60.7\pm1.5$ & $1.35\pm0.10$ & $\textless0.9\%$ & $67.0\pm3.0$ &$1.90\pm0.20$ & $\textless0.6\%$  \\
		\texttt{I2G}~\cite{Liu2022} & $59.8\pm1.3$ & $1.30\pm0.07$ & $\textless0.03\%$ & $71.0\pm1.4$ &$1.64 \pm 0.10$ & $\textless0.01\%$  \\
		\texttt{PR++}~\cite{Dual} & $61.1\pm1.4$ & $1.34\pm0.10$ & $\textless0.5\%$ &$69.5\pm2.2$&$1.76\pm0.17$& $\textless0.2\%$   \\
		\texttt{XM}~\cite{shi2022xmorpher} & $53.6\pm1.5$ & $1.46\pm0.09$ & $\textless1\%$ &$66.3\pm2.0$&$1.92\pm0.15$& $\textless0.1\%$   \\
		\texttt{DMR}~\cite{Chen2022} & $60.6\pm1.4$ & $1.34\pm0.09$ & $\textless0.7\%$ &$69.2\pm2.4$&$1.79\pm0.18$& $\textless0.4\%$   \\
		\texttt{Ours} & \boldmath{$62.8\pm1.2$} & \boldmath{$1.22\pm0.07$} & $\textless0.03\%$ & \boldmath{$72.1 \pm1.4$}&\boldmath{$1.58 \pm 0.11$}  & $\textless0.007\%$  \\
		\bottomrule
	\end{tabular}
\end{table}

\begin{figure}[t]
	\centering
	\includegraphics[width=1.0\columnwidth]{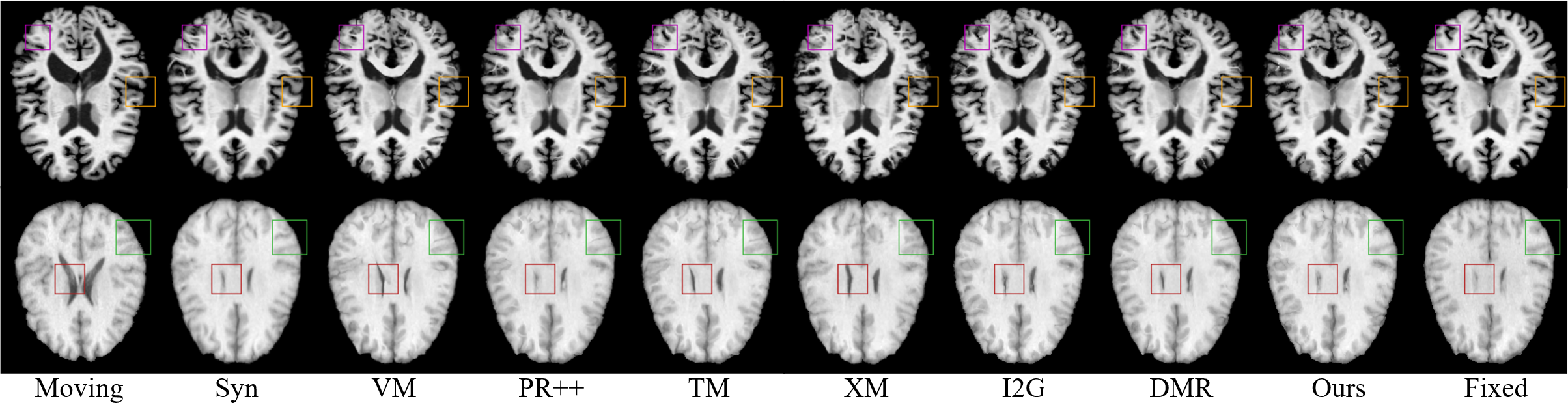}
	\caption{Visualized registration results from different methods on Mindboggle (top row) and LPBA (bottom row).}
	\label{f:results}
\end{figure}

\subsubsection{Implementation Details.}
Our method was implemented with PyTorch, using a GPU of NVIDIA Tesla V100 with 32GB memory.
The regularization term $\lambda$ and neighborhood size $n$ were set as $1$ and $3$.
For the encoder part, we used the same convolution structure as~\cite{Liu2022}.
In the pyramid decoder, from coarse to fine, the number of attention heads were set as $8,4,2,1,1$, respectively.
We used 6 channels for each attention head.
The Adam optimizer~\cite{kingma2014adam} with a learning rate decay strategy was employed as follows:
\begin{equation}
	lr_m =  lr_{init}\cdot(1-\frac{m-1}{M})^{0.9}, m = 1, 2, ... ,M
\end{equation}
where $lr_m$ represents the learning rate of $m$-th epoch and $lr_{init}=1$ represents the learning rate of initial epoch.
We set the batch size as $1$, $M$ as $30$ for training.

\subsubsection{Comparison Methods.}
We compared our method with several state-of-the-art registration methods:
(1)~\texttt{SyN}~\cite{AVANTS2008}: a classical traditional approach, using the $SyNOnly$ setting in ANTS.
(2)~\texttt{VoxelMorph(VM)}~\cite{VoxelMorph}: a popular single-stage registration network.
(3)~\texttt{TransMorph(TM)}~\cite{Chen2022a}: a single-stage registration network with SwinTransformer enhanced encoder.
(4)~\texttt{PR++}~\cite{Dual}: a pyramid registration network using 3D correlation layer.
(5)~\texttt{XMorpher(XM)}~\cite{shi2022xmorpher}: a registration network using CAT modules for each level of encoder and decoder.
(6)~\texttt{Im2grid(I2G)}~\cite{Liu2022}: a pyramid network using a coordinate translator.
(7)~\texttt{DMR}~\cite{Chen2022}: a registration network using a Deformer and a multi-resolution refinement module.

\subsubsection{Quantitative and Qualitative Analysis.}
The numerical results of different methods on datasets Mindboggle and LPBA are reported in Table~\ref{tab1}.
It can be observed that our method consistently attained the best registration accuracy with respect to DSC and ASSD metrics.
For the DSC results, our method surpassed the second-best networks by $1.7\%$ and $1.1\%$ on Mindboggle and LPBA, respectively.
We further investigated the statistical significance of our method over comparison methods on DSC and ASSD metrics, by conducting the paired and two-sided Wilcoxon signed-rank test.
The null hypotheses for all pairs (our method $v.s.$ other method) were not accepted at the 0.05 level.
As a result, our method can be regarded as significantly better than all comparison methods on DSC and ASSD metrics.
Table~\ref{tab1} also lists the percentage of voxels with non-positive Jacobian determinant ($ \% |J_{\phi}|\le0 $).
Our method achieved satisfactory performance, which was the best among all deep-learning-based networks.

\begin{figure}[t]
	\centering
	\includegraphics[width=1.0\columnwidth]{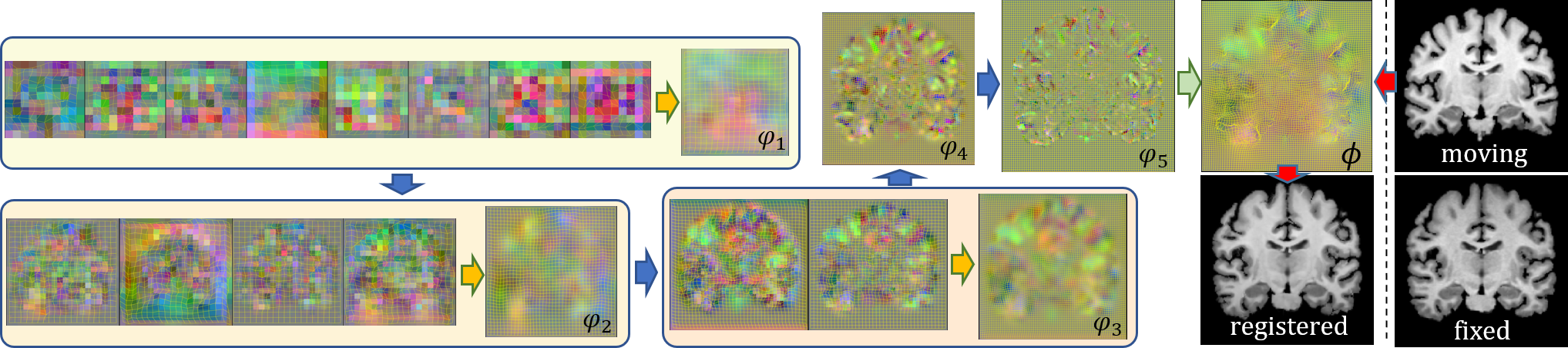}
	\caption{Visualization of the generated multi-level deformation fields ($\varphi_1$-$\varphi_5$) to register one image pair. At low-resolution levels, multiple deformation sub-fields are decomposed to effectively model different motion modalities.}
	\label{f:deformation}
\end{figure}

Fig.~\ref{f:results} visualizes the registered images from different methods on two datasets.
Our method generated more accurate registered images, and internal structures can be consistently preserved using our method.
Fig.~\ref{f:deformation} takes the registration of one image pair as an example to show the multi-level deformation fields generated by our method.
Our ModeT effectively modeled multiple motion modalities and our CWM fused them together at low-resolution levels.
The final deformation field $\phi$ accurately warped the moving image to registered with the fixed image.

\section{Conclusion}
We present a motion decomposition Transformer (ModeT) to naturally model the correspondence between images and convert this into the deformation field, which improves the interpretability of the deep-learning-based registration network.
The proposed ModeT employs the multi-head neighborhood attention mechanism to identify various motion patterns of a voxel in the low-resolution feature map.
Then with the help of competitive weighting module and pyramid structure, the motion modes contained in a voxel can be gradually fused and determined in the coarse-to-fine pyramid decoder.
The experimental results have proven the superior performance of the proposed method.

\section*{Acknowledgements}
This work was supported in part by the National Natural Science Foundation of China under Grants 62071305, 61701312, 81971631 and 62171290,
in part by the Guangdong Basic and Applied Basic Research Foundation under Grant 2022A1515011241,
and in part by the Shenzhen Science and Technology Program (No. SGDX 20201103095613036).

%
%
%
 \bibliographystyle{splncs04}

\bibliography{bib2}
\end{document}